# Improving Fuzzy-Logic based Map-Matching Method with Trajectory Stay-Point Detection

Minoo Jafarlou*, Omid Mahdi Ebadati E., and Hassan Naderi

*Abstract*—**The requirement to trace and process moving objects in the contemporary era gradually increases since numerous applications quickly demand precise moving object locations. The Map-matching method is employed as a preprocessing technique, which matches a moving object point on a corresponding road. However, most of the GPS trajectory datasets include stay-points irregularity, which makes map-matching algorithms mismatch trajectories to irrelevant streets. Therefore, determining the stay-point region in GPS trajectory datasets result in better accurate matching and more rapid approaches. In this work, we cluster stay-points in a trajectory dataset with DBSCAN and eliminate redundant data to improve the efficiency of the map-matching algorithm by lowering processing time. We reckoned our proposed method's performance and exactness with a ground truth dataset compared to a fuzzy-logic based map-matching algorithm. Fortunately, our approach yields 27.39% data size reduction and 8.9% processing time reduction with the same accurate results as the previous fuzzy-logic based map-matching approach.**

*Index Terms*—**DBSCAN, Fuzzy-logic, GPS Trajectory, Map-matching, Stay-point**

## I. INTRODUCTION

A wide range of management plans and intelligent transportation systems require fast and accurate location information, such as urban traffic management, navigation systems, accident management, and emergency responses. Global Positioning System (GPS) is the most common locator device in urban areas. GPS trajectory data should be matched to the road network using Map-matching algorithms to analyze and use the vehicle trajectories in the spatial road network. This process assembles the raw GPS data into selected road segments [1]. Unfortunately, GPS devices may malfunction in some cases. In dense urban environments, where tall buildings exist, it will be challenging for the satellites to locate moving objects. Therefore, in such environments, these devices' location has never been entirely accurate, and they report the position of the moving objects with some error [2]. Map-matching uses these noisy inputs and mismatches the point to the wrong roads. To solve this problem, we wipe the unnecessary noisy inputs from the map-matching process; thus, we achieve more accurate results faster. We aim to increase the efficiency of this algorithm by reducing the processing time.

*M. Jafarlu., Researcher, Department of Knowledge Engineering and Decision Science, Kharazmi University (correspondence to mijafarlou@gmail.com)

O.M. Ebadati. E., Associate Professor, Department of Knowledge Engineering and Decision Science, Kharazmi University

H. Naderi., Associate Professor, Department of Computer Engineering, University of Science and Technology (IUST)

Map-matching algorithms are mainly used in navigation, mapping, and tracking. In most cases, navigation algorithms use online data; however, tracking and mapping methods usually work with offline data [3]. Map-matching algorithms are split into four leading groups: 1-Geometrical map-matching algorithms, 2-topological map-matching algorithms, 3-probabilistic map-matching algorithms, and 4-advanced map-matching algorithms [4]. Advanced map-matching algorithms predominantly use fuzzy-logic, Hidden Markov Model (HMM) algorithms, or Kalman filters. These algorithms estimate the GPS points' location more accurately; nonetheless, attaining this exactness costs a high processing duration. Viewing this problem, we choose to work with a fuzzy-logic based map-matching algorithm due to its precision in matching high-frequency sampling GPS points.

Map-matching associates GPS data to an interconnected avenue and enhances the accuracy of urban streets trajectories by matching the movement points to suitable road networks, despite some problems. Abnormalities and errors in GPS trajectory may cause inaccurate results in a map-matching process. One of the most critical problems with raw GPS trajectory datasets is the stay-points problem. This problem imports many GPS points into the dataset that are unfavorable in the map-matching process. Besides, processing all of these points in the map-matching algorithm increases the algorithm's processing time. Noted stationary states need to be detected in a dataset for two main reasons. First, the processing time is crucial in analyzing, evaluating, and comparing a method with other methods. Second, these points may lead to a disturbance in the map-matching process. Map-matching algorithms mostly use the direction of a vehicle and the distances as inputs. Stay-point regions have a drastic effect on these two parameters. If these parameters are inaccurate, the map-matching algorithm may match the points to unrealistic roads. This mismatching could even change the points' direction up to 180 degrees and create matching problems. This research aims to use the density-based spatial clustering of applications with noise (DBSCAN) clustering method to detect stay-points and remove the inessential parts of the dataset which are not favorable in the map-matching process. In this approach, we hope to enhance the fuzzy-logic algorithm's performance by eliminating these data and reducing the dataset size and processing duration. In addition, by reducing the dataset size, we can decrease large-scale storehouse capacity and computation time.

This research improves the map-matching algorithm by preprocessing the raw GPS trajectory to identify stay-point regions and eliminate surplus points in data. The paper's noteworthy contributions are reducing processing time without affecting the accuracy of correct link identification and data reduction.

The rest of the paper is organized as follows: In division 2, we introduce preliminaries and algorithms used in section 3. Next, in branch 3, we offer our solution. In section 4, we argue the result of the proposed method. Later, in part 5, affiliated works of existing map-matching algorithms and study areas are presented. Finally, the conclusion and future work are declared.

## II. PRELIMINARIES

### *A. Overview of input data*

1. Trajectory: A sequence of GPS points that records the motions of a moving object over a specific period [$tbegin$, $tend$] is called the trajectory of that moving object.

2. Digital Road network: Every road network contains topological and geometric information. A road network can mainly be divided into intersections, squares, urban roads, rural roads, and highways. A digital road network is a road network in a graph, similar to an actual network map, implemented in a computer environment.

### *B. Map-matching process*

This process needs a digital road network and GPS trajectory as inputs. The digital road network is a simulation of the road system shown with arcs and links. Like a real road network, this stimulation consists of many resembled roads, junctions, and dead ends [5]. The process of mapping GPS points on the links of the digital road network is called map-matching [4].

A vehicle is moving along a finite system of roads, $\bar{\mathcal{N}}$. We do not precisely know the road system, $\bar{\mathcal{N}}$, instead, we have a network representation, $\boldsymbol{\mathcal{N}}$, consisting of a set of curves in $\mathbb{R}^2$. It is assumed that a one-by-one correspondence exists between the roads in $\bar{\mathcal{N}}$ and the arcs in $\boldsymbol{\mathcal{N}}$. Arcs are assumed to be a piece-wise liner, representing single roads. Therefore, arc A$\epsilon\boldsymbol{\mathcal{N}}$ can be defined by a finite sequence of points, $(A_0, A_1, \dots, A_{n_A})$. In fact, arc A consists of these points, which are endpoints of the line segment. The first point and the last point are usually called nodes. These are the endpoints of an arc, making it possible to move from one arc to another. Hence, they correspond to dead-ends or intersections in the road system. Estimation of the vehicle's location at time t is given. $\mathcal{P}_t$ represents this estimation. $\bar{\mathcal{P}}_t$ signifies the vehicle's actual location at time t. Concerning these, the map-matching algorithm has two goals. First, to specify the street $\bar{A}\epsilon\bar{\mathcal{N}}$ correspondence to the vehicle's actual location $\bar{\mathcal{P}}_t$ which is acquired by matching the estimated location $\mathcal{P}_t$ with an arc A$\epsilon\boldsymbol{\mathcal{N}}$. Second, to specify the position on A that best corresponds to $\bar{\mathcal{P}}_t$ [5, 6].

The two fixed point and matched point concepts have been used in this research. Fixed points indicate that map-matching processing is not applied to them, and they are an estimated location of moving objects. However, the matched points are the ones that the matching process was performed on them.

Evaluation method: There are two standard criteria for evaluating the map-matching method's efficiency: processing time or the method and the number of correct matches and wrong matched roads.

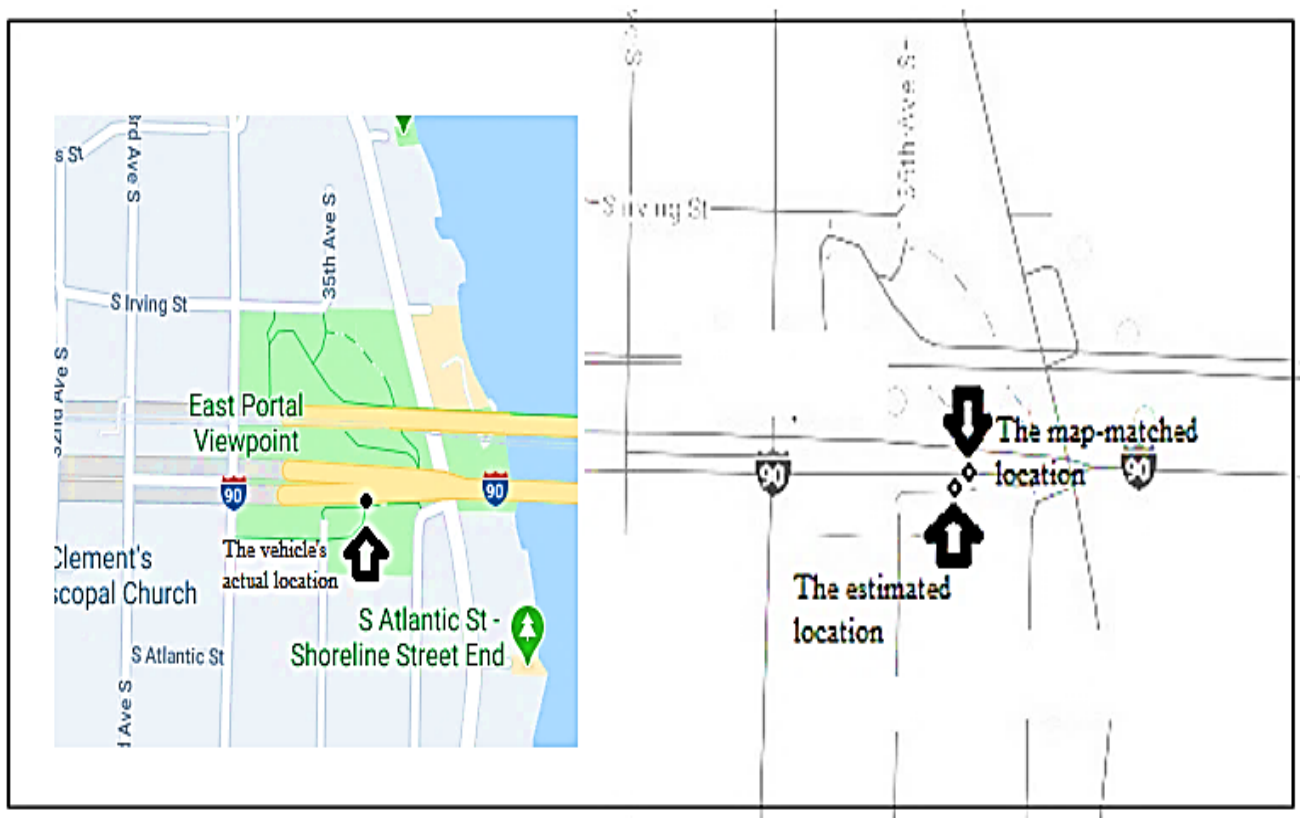

Fig. 1. The map-matching process

### *C. Fuzzy-logic system*

Fuzzy logic is a many-valued logic in which the variables' values are between 0 and 1, both inclusive for each number. It has three main steps: 1. Fuzzification means fuzzifying all "crisp" input values with membership functions to a fuzzy input set. 2. Interface engine: Operating all applicable rules in the "fuzzy rules base" to calculate fuzzy output functions. 3. Defuzzification: De-fuzzifying fuzzy output set to "crisp" output values. Fig. 2. shows this process.

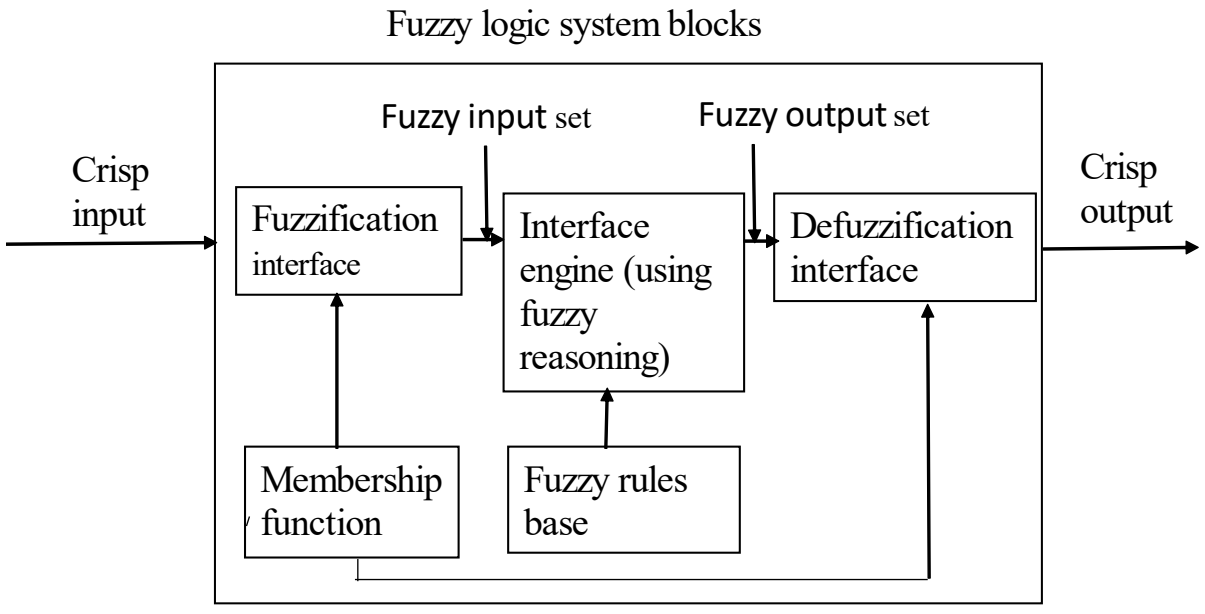

Fig. 2. Fuzzy-logic system blocks

### *D. Fuzzy-logic based map-matching algorithm overview*

The fuzzy-logic based map-matching process has two steps: finding the correct link the vehicle is traveling on and tracking the vehicle along with the link [7]. Some methods [8], [9] divide the first step into two sub-steps: Finding an initial correct link and finding a correct link when the vehicle crosses an intersection. Fig. 6. illustrates this process, and algorithm 1 shows the pseudo-code of this process.

1. Initial map-matching process (IMP): Selecting the initial link to assign the initial position is known as the initial map-matching process (IMP). The IMP uses the two variables as inputs of IMP: Fixed point's perpendicular distance to the link (PD) and the directional difference between the vehicle's motion and the link direction known as heading error (HE).

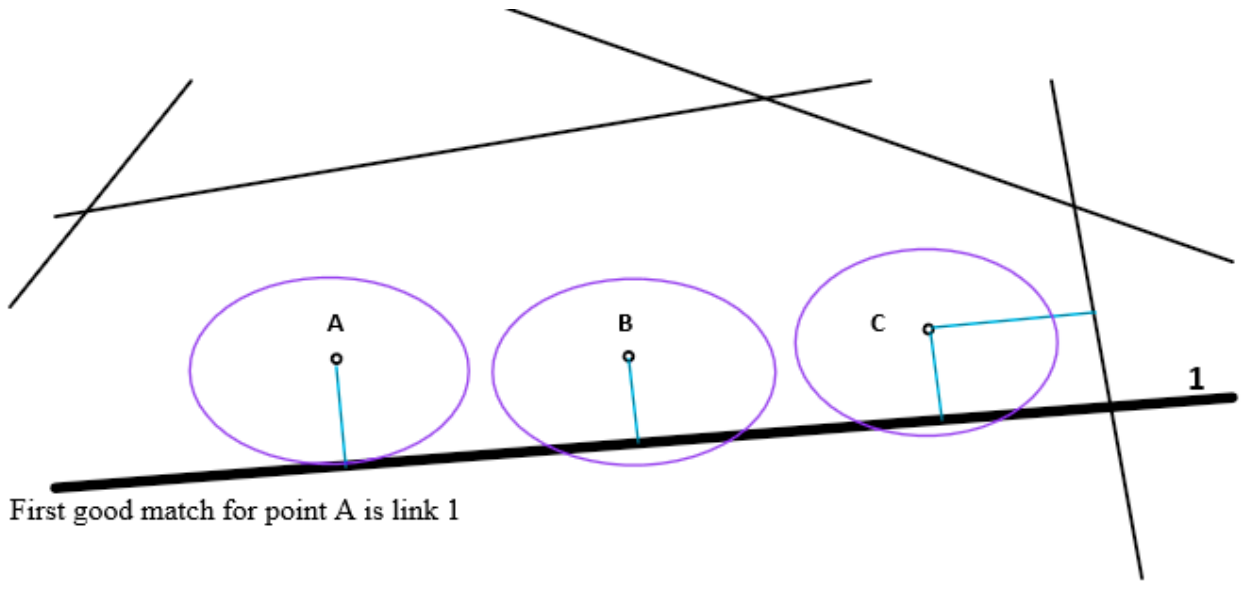

Fig. 3. Initial map-matching

After IMP, we need a subsequent map-matching process (SMP). The SMP is a process to match the fixed points in the sequence of IMP. Two models of SMP are proposed: 1. SMP along with the link (SMP-1), 2. SMP at the intersection (SMP-2). SMP-1 aims to map-match the succeeding points with the link identified by the IMP or SMP2. SMP-2 detects a new link for the first point among volunteer links when the vehicle crosses an intersection. After SMP-2 detects a new link, SMP-1 resumes matching fixed points following the new link.

2. Subsequent map-matching process along with the Link (SMP-1): SMP-1 determines whether a subsequent fixed-point corresponds to a previously selected link. Fig. 4. shows A, as the previous map-matched point and B, as the current fixed point. The task of SMP-1 is to select the correct link for the following fixed point (B). Expression d refers to the distance between the last matching point on the map and the intersection, and d2 refers to the distance traveled by the vehicle during the last period. The distance between these two distances ($\Delta d = (d-d2)$) can be used to observe whether the vehicle is crossing the intersection or not. For example, if $\Delta d$ is negative, it is most likely that the vehicle has already crossed the intersection. Both $\propto$ and $\beta$ define the position of the fixed-point B in connection with link 1. If both of these angles are less than 90 degrees, it is more likely that the vehicle did not cross the intersection. The angles $\theta$ and $\theta'$ are the vehicle's direction indicators at B and A, respectively. The absolute value difference between these two angles $abs(\theta - \theta')$ provides B's heading increase (HI) in the last period. Therefore, the FIS fuzzy variables are 1. Vehicle speed (v) 2. heading increase (HI), 3. distance traveled by the vehicle during the last period ($\Delta d$) 4. $\propto$, and 5. $\beta$. This FIS's output is a probability of matching the new fixed point with the same link as the formerly fixed point.

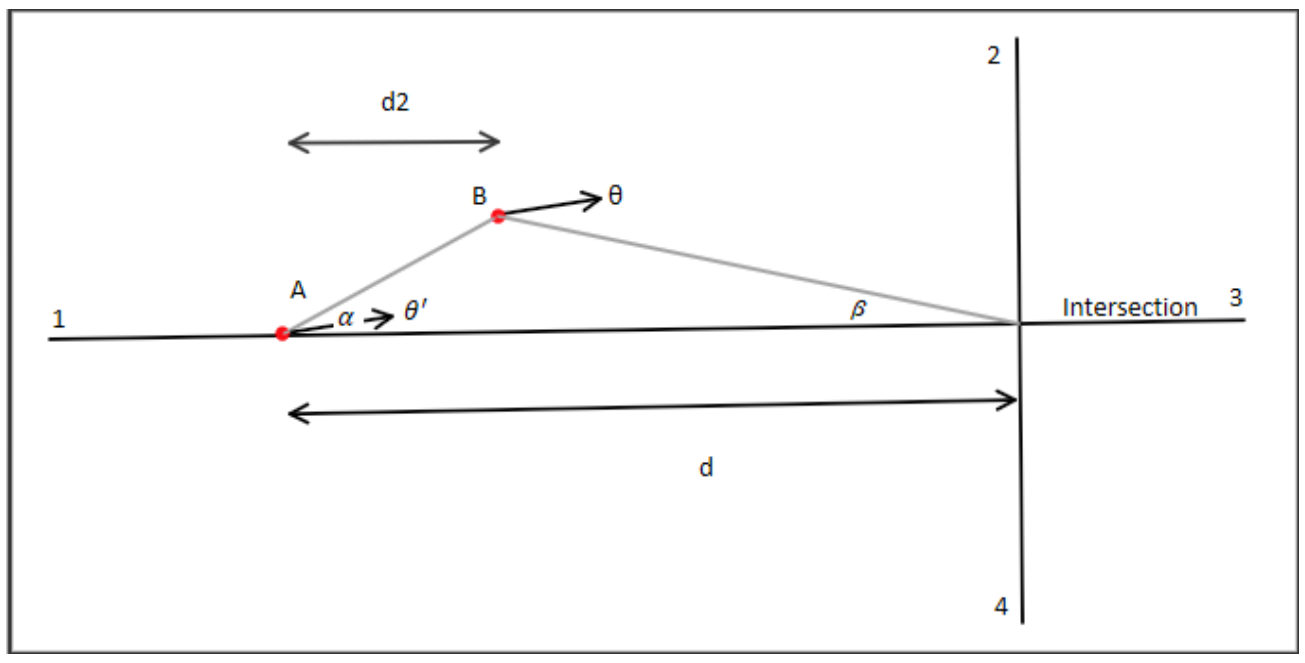

Fig. 4. $\theta'$ is the vehicle's direction at previous point. $\Delta d$ is $d - d1$. Abs $(\theta - \theta')$ is heading increment.

3. Subsequent map-matching process at an Intersection (SMP-2): SMP-2 starts when the vehicle passes the intersection. Like IMP, the function's inputs are fixed points' a perpendicular distance (PD) and heading error (HE). Besides, link connections and distance errors input variables are used in this FIS. The device is moving on link 1. Then, the last map-matched point is on link number 1. SMP-2 selects a new link for map-matching point B. Volunteer links for this point are 2, 3, and 4. Since the previous point's position is on link 1, the link connection is an essential criterion for identifying the correct link. The expression d refers to the distance traveled by the vehicle during the last period. If the device is on links 2, 3, and 4, then d2, d3, and d4 represent the shortest path traveled by the vehicle. The distance error is the difference between d and d2 or d3 or d4. The link that provides the least distance error is a strong candidate for the correct link.

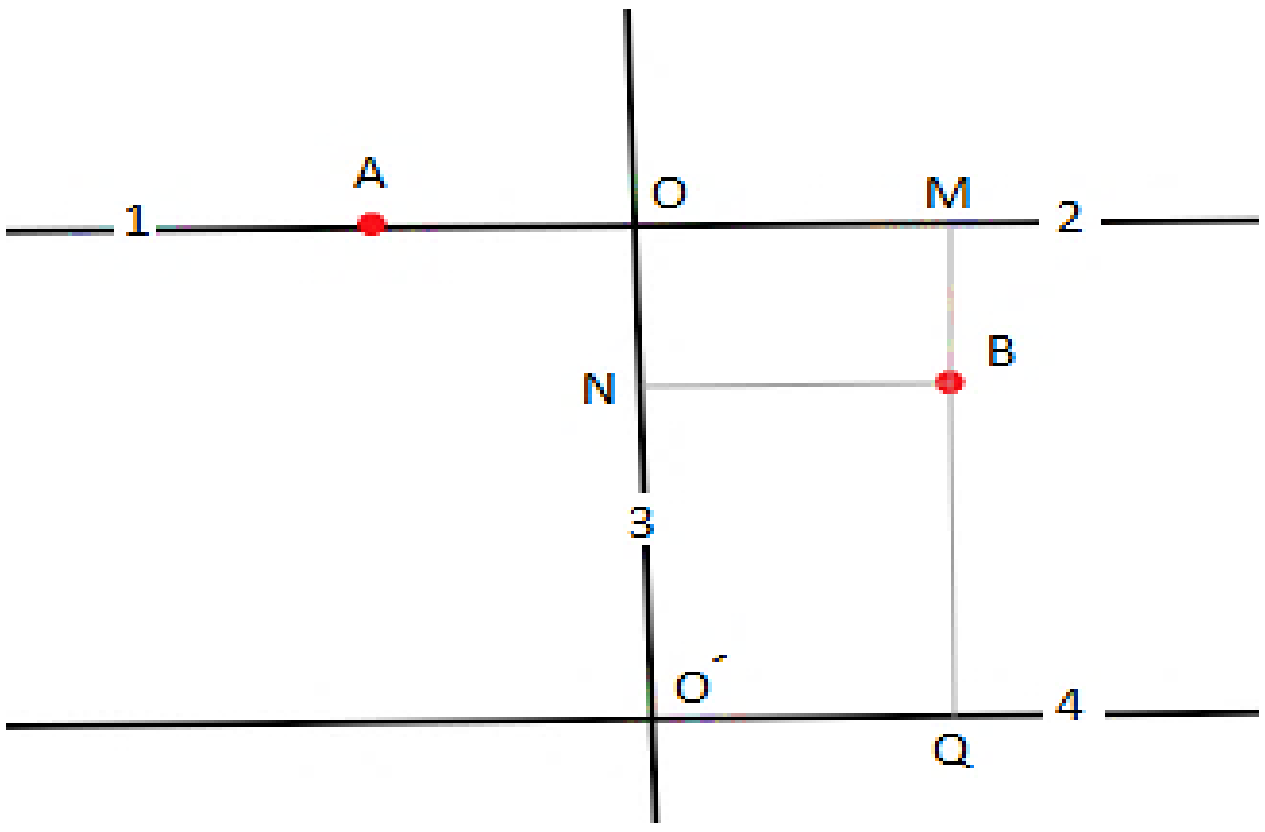

Fig. 5. d is the traveled distance by vehicle, d2= AO+OM, d3= AO+ON, d4= AO+OO'+OQ

**Algorithm 1**: Fuzzy logic map-matching algorithm

**Input:** a set of GPS Trajectory *trajectory*=$\{p_1, p_2, ..., p_m\}$, Digital Road Network *roadnetwork*= $\{(e_1, (v_1, v_2), l_1), (e_2, (v_2, v_3), l_2), ..., (e_n, (v_n, v_q), l_r\}$

**Output:** Map-matched Trajectory *matchedtajectory*$\{(p_1, e_1), ...., (p_m, e_z)\}$

*trajectory* ← Convertto-SpatialObject (*trajectory*);
*roads* ← Create-DigitalRoadNetwork (*roadnetwork*);
*list* ← Initial-MapMatching (*trajectory, roads*);
*edited-trajectory* ← *list.trajectory*;
*point-index* ← *list.index*;
*current-link* ← *list.currentlink*;
**for** *j in point-index* **do**
  *predicted-value* ← Subsequent-MapMatching-1 (*edited-trajectory, roads, current-link, j*);
  **if** *predicted-value* ≥ *60* **then**
    *edited-trajecory.EdgeID[j]* ← *edited-trajectory.EdgeID[j-1];*
  **else**
    *current-link* ← Subsequent-MapMatching-2 (*edited-trajectory, roads, current-link, j*);
    *edited-trajecory.EdgeID* ← *current-link.EdgeID*;
  **end if**
**end for**
*matchedtajectory* ← SpatialPointDataFrame (*edited-trajectory*);
**return** *matchedtajectory*

Algorithm. 1. Pseudo-code for fuzzy map-matching algorithm

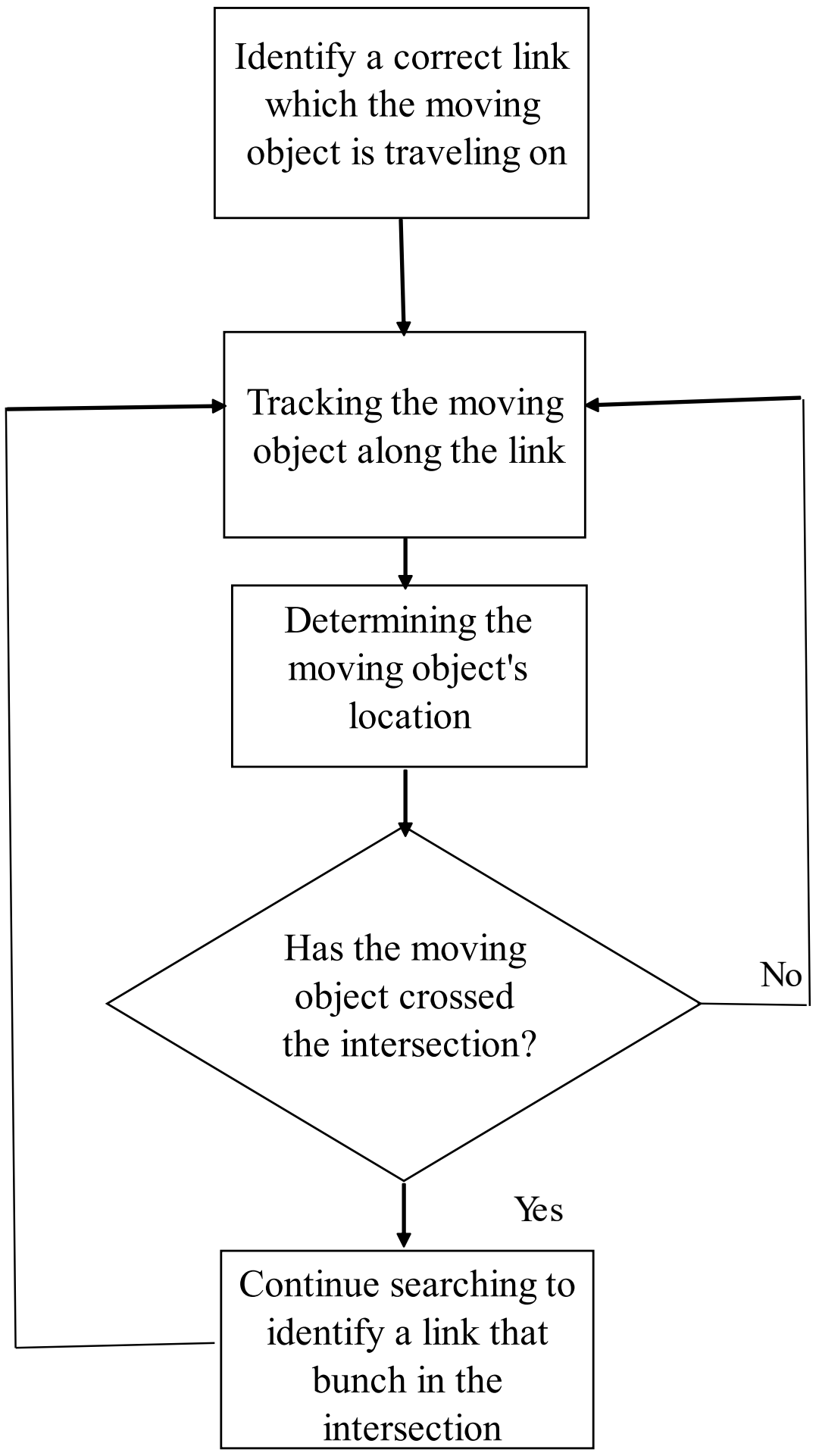

Fig. 6. Map-matching process

*E. DBSCAN algorithm*

In 1996, Ester et al. introduced the density-based clustering algorithm DBSCAN. They designed this algorithm to determine clusters and noises in the spatial datasets [10] . The two following parameters were used in this method to shape a dense area: 1) Eps-neighborhood (Eps) and 2) the minimum number of points (*MinPt*) [11]. For each point, the neighboring points are defined as points that exist within a radius of Eps [12]. DBSCAN could detect arbitrary shape clusters and also could distinguish noises. Also, DBSCAN effectively works with large spatial databases [13, 14].

This algorithm begins with an optional point that has not yet been visited. The point's neighbors in a radius of Eps are retrieved. If these neighboring points exceed *MinPt*, a cluster is created, and that specific point is marked as the core point. Otherwise, the point is identified as noise unless it is found in a radius of other core points and becomes part of their cluster. Adding the reachable points to the cluster continues until the complete cluster forms. Then, new unvisited points are processed to be detected as noise or part of a cluster. The cluster would continue growing by adding reachable points from the core point [15].

Determining the values for the parameters is a crucial part of solving a clustering problem. Broadly, each parameter affects the creation of the cluster in a wide range. It is essential to plot the dataset's kth nearest neighbor to determine the optimal value for the *Esp* parameter in the DBSCAN clustering algorithm. Determining *MinPt* for the DBSCAN algorithm is often extracted from the problem definition.

The plot computes the *kth* nearest neighbor for each point in the dataset and arranges them in ascending order. This plot shows us a variety of density levels in the dataset. The plot will have one sharp spot if the dataset has low-density distribution. Likewise, a sharp spot in this plot "elbow-shaped" presents good values for the Eps parameter. However, if the dataset has a great variety of densities, the curve would have multiple sharp spots [10]. Substantial deviations from the smooth path of the curves were observed. It is essential to mention that each curve, corresponding with the connecting noise point, represents a density level. In other words, the elbow method is used to find the optimal amount of epsilon. The elbow of the diagram corresponds to the threshold for sudden changes in the diagram.

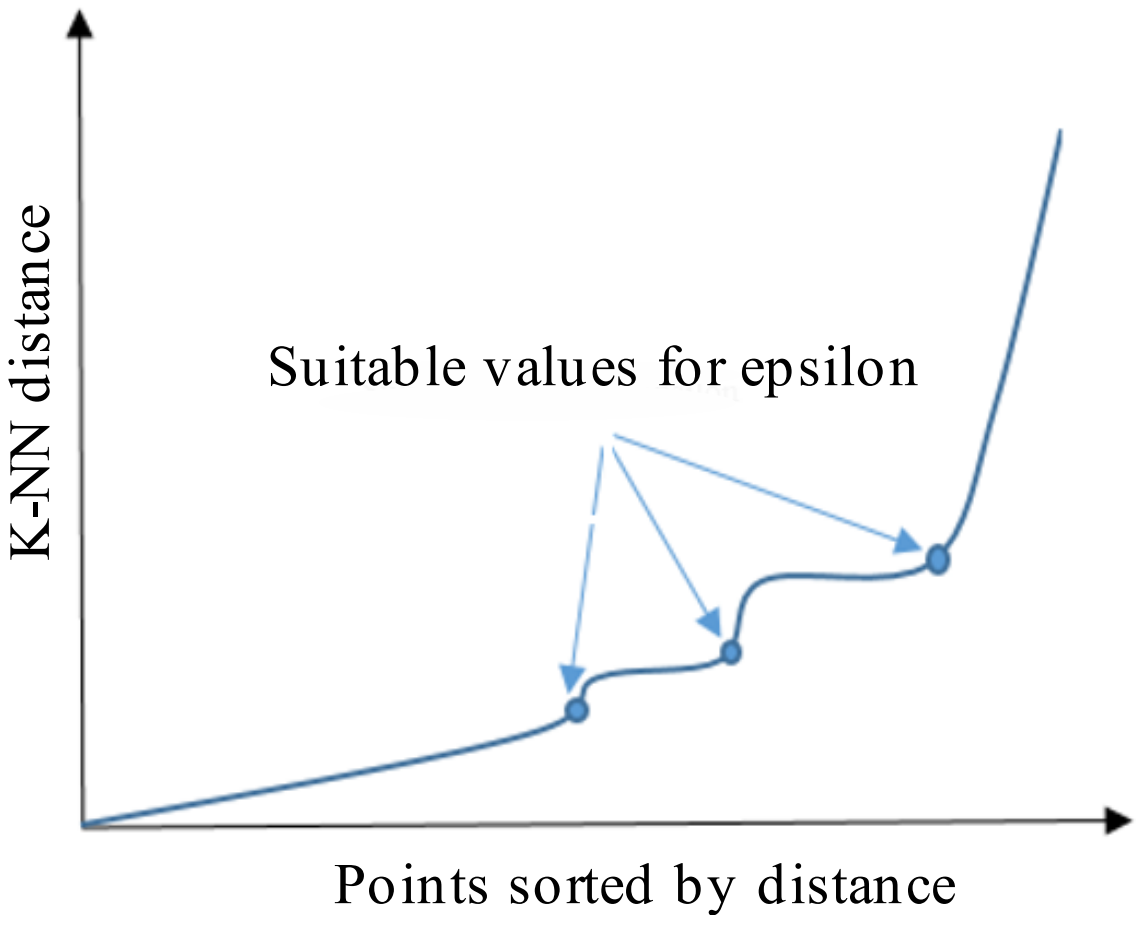

Fig. 7. Different density levels

Fig. 7. presents a plot with three-level densities. Each sharp spot stands for a different density level. As seen in the diagram, there are three sharp changes in the curve. These three sharp changes mean there are three suitable values for Eps parameters [10].

*F. Stay-point problem definition*

Raw trajectory datasets have various primary noises. The most important one in these datasets is the stay-point problem. In these cases, the navigation sensor misses the satellite and reports unrealistic locations of an immobile moving object. It seems that the moving object is moving in a strange and confusing path instead of staying in a fixed coordinate. Fig. 8., and Fig. 9. Illustrate stay-points regions. This abnormality in the dataset can mislead a map-matching algorithm and prevent it from reaching an accurate result. The mismatching problem is evident in Fig. 10. Even if it does not miss the satellite, it will send the previous location signal to the satellite and increase the dataset volume. In the end, it will damage the accuracy and the processing time of the map-matching algorithm.

A stay-point region is a dense part of the trajectory where the moving object is stationary for a while [1]. Two parameters are used to identify stationary points: A time threshold ($\tau$) and distance threshold ($\delta$) [1]. In a specified trajectory where each point in the track contains a timestamp and location, Track= <$p_1, p_2, \dots, p_n$>. Stay-point is defined as a sub-trajectory <$p_i, \dots, p_j$>, that $\forall k \epsilon [i,j), Dist(p_k, p_{k+1}) < \delta, Int(p_i, p_j) > \tau$. Thus, $= (x, y, t_a, t_l)$ , where

$$s.x = \sum_{k=i}^{j} p_k.x/|s| \qquad (1)$$

$$s.y = \sum_{k=i}^{j} p_k.y/|s| \qquad (2)$$

(1) and (2) equations stand for the average x and y coordinates of the stay-point $s$; $s.t_a = p_i.t$ is the moving object's arrival time on $s$ and $s.t_l = p_i.t$ represents the moving object's leaving time[1] , [16].

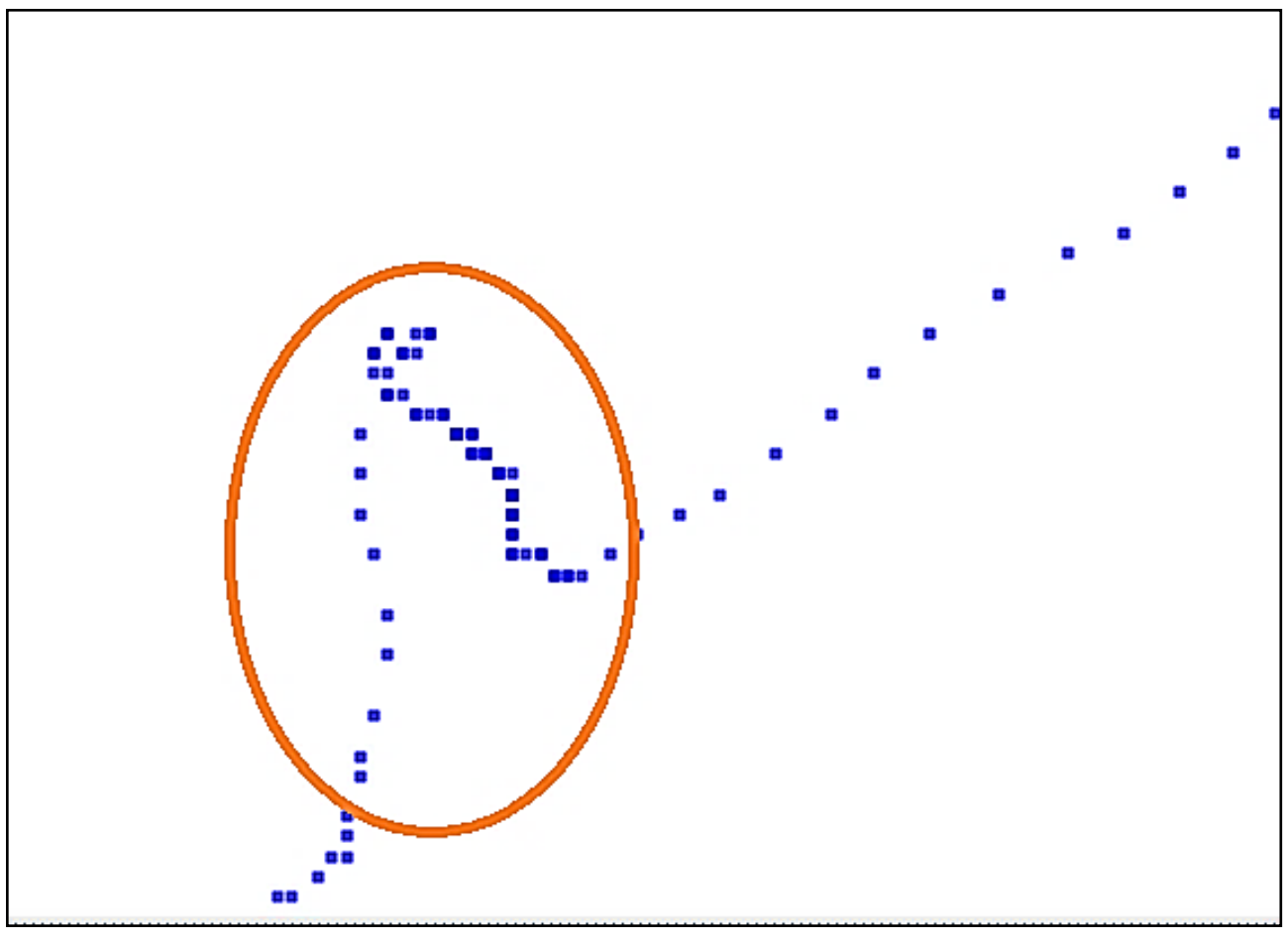

Fig. 8. Sample of stay-point region on the dataset

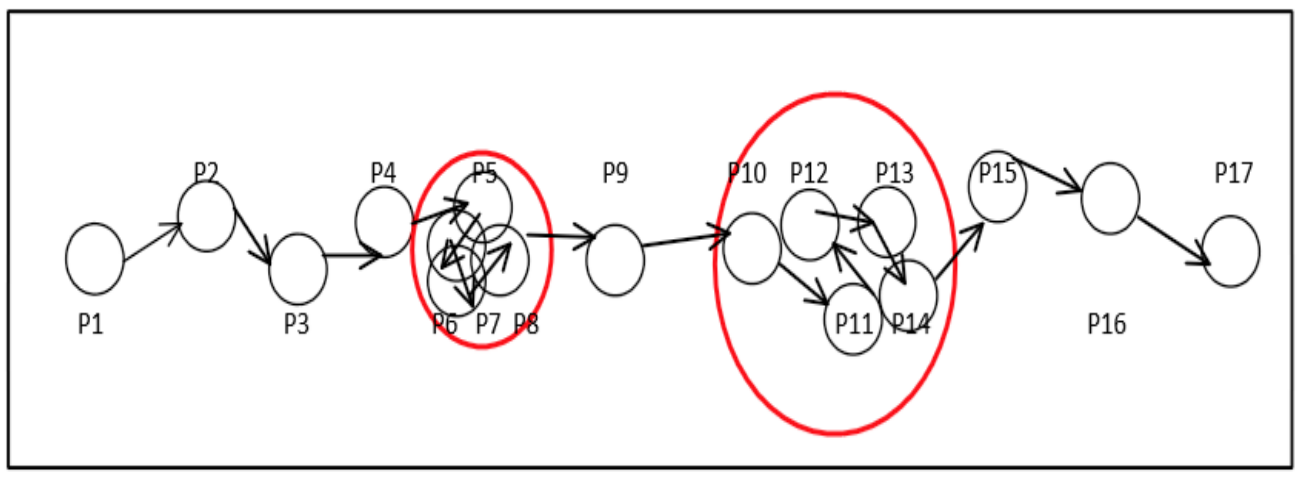

Fig. 9. Definition of stay-point regions on a trajectory

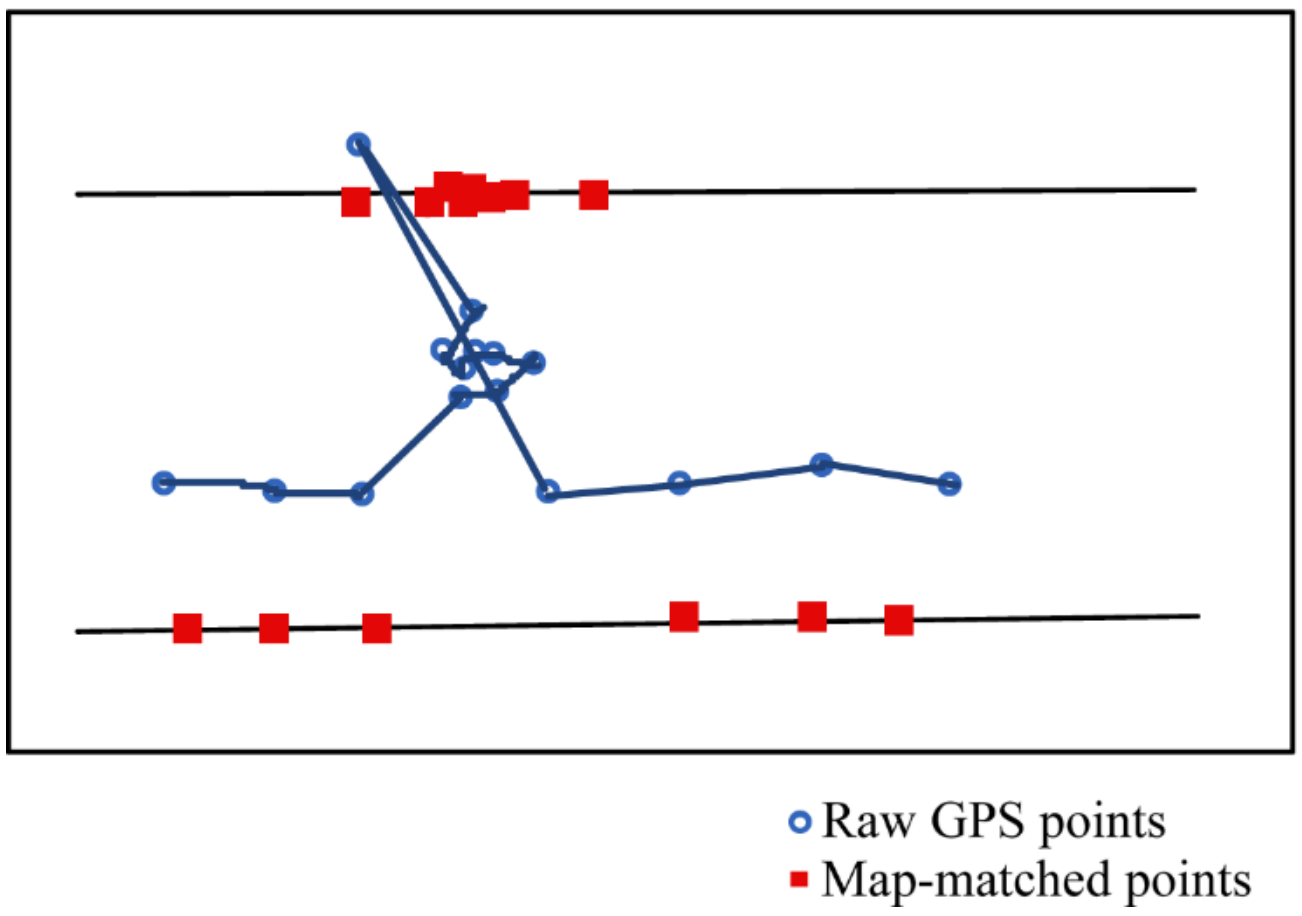

Fig. 10. Stay-point region misleads the map-matching process. Blue dots are raw trajectory, red dots are misled and mismatched GPS points to another road link

## III. PROPOSED METHOD

As cited in the introduction, stay-points misleads map-matching methods, increase the unneeded computation duration of the procedure, and raise mismatches; the key is to cluster this aberration in the trajectory. In this section, we first define the ground truth real-world data, then introduce our proposed method.

### *A. Consumption data*

The map-matching algorithm inputs are GPS trajectory and topological and geometric information of the road network. We use these two inputs to determine the roads on which the vehicle passes.

1. Digital road network data: A digital road network is a computer simulation graph of the city's roads network. This information is used to determine the direction and curve of the roads. Input data consists of edge id, line string, and node id. Line string is multiple connected lines representing a road –each edge id represents a road section. Each road is determined by a segment of lines, called "link." Fuzzy-logic map-matching algorithm will identify the corresponding link for each point. Then, output data is a string of associating links.

2. GPS trajectory: The GPS trajectory [17] is collected from the movement of a vehicle on the roads of Seattle with a 7351 GPS record. This dataset includes latitude and longitude with a time stamp. The data are recorded in an 80 km trip with a data transmission frequency of 1 Hz. This data is used to test map-matching algorithms, evaluate each method's efficiency, and compare the techniques with each other.

Seattle's GPS trajectory dataset [17] and the corresponding digital road network are shown in Fig. 11. This ground truth dataset is designed for testing map-matching algorithms.

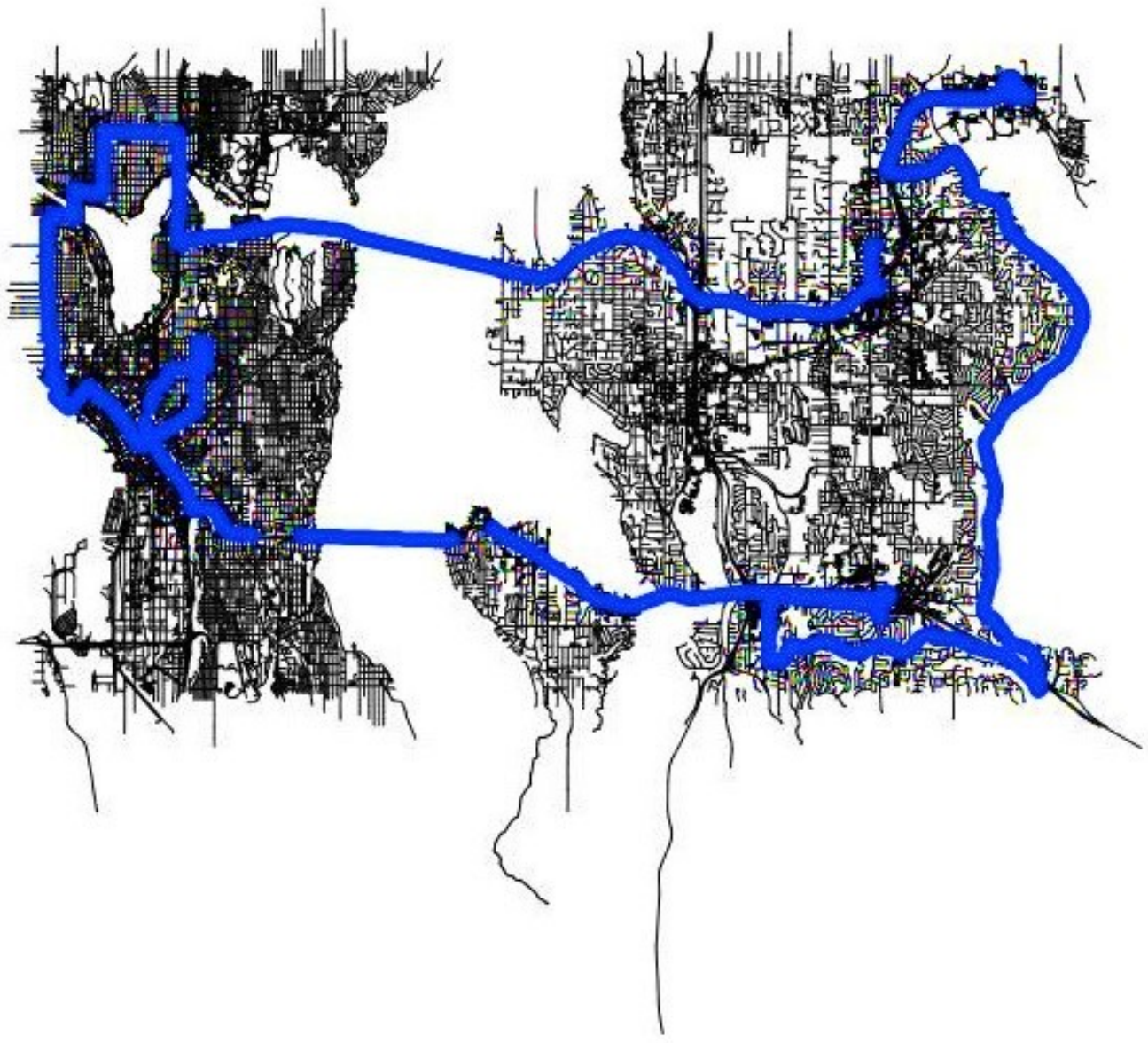

Fig. 11. Seattle's digital road network tested GPS trajectory, which corresponds to the limited geographical area of the data. This road network consists of more than 8500 roads.

3. Ground truth data: Data also contains a true path as "edge id"s that vehicle encountered during the test trip. The valid route includes 399 road edges.

### *B. Proposed method*

The primary purpose of this research is to cluster the raw GPS trajectory dataset with the DBSCAN algorithm to detect the stay-point part and eliminate extra points in data before importing the data into the fuzzy-logic based on the map-matching algorithm.

Nevertheless, other approaches such as specified time span, the maximum number of points per specific distance thresholds, and standard clustering methods might be used to discover and reduce stay-points. Although these approaches may help us get closer to the consequence, none are satisfactory outcomes. When the satellite misses the GPS receiver, it cannot get accurate data from the receiver, and it would yield inaccurate results in the familiar approaches mentioned above. In addition, when points are close to each other, for example, when the vehicle is in a traffic jam. Considering a threshold to delete data points might lead to eliminating valuable data necessary for the map-matching process. On the other hand, other clustering algorithms do not work as well as DBSCAN for trajectory clustering due DBSCAN produces the most accurate result among existing methods. As a matter of fact, DBSCAN defines clusters based on the local density of the data element in big spatial datasets. Moreover, it is robust to outliers and does not require the number of clusters predefined. Choosing DBSCAN for clustering data is paramount since points in fewer dense areas separate the clusters in DBSCAN. In other words, lower density zones split the regions. This component makes DBSCAN less resistant to noise and clusters shapeless trajectories more efficiently. DBSCAN clustering solves our problem by clustering the unnecessary data and reducing the size and time. The architecture of the proposed procedure is depicted in Fig. 12.

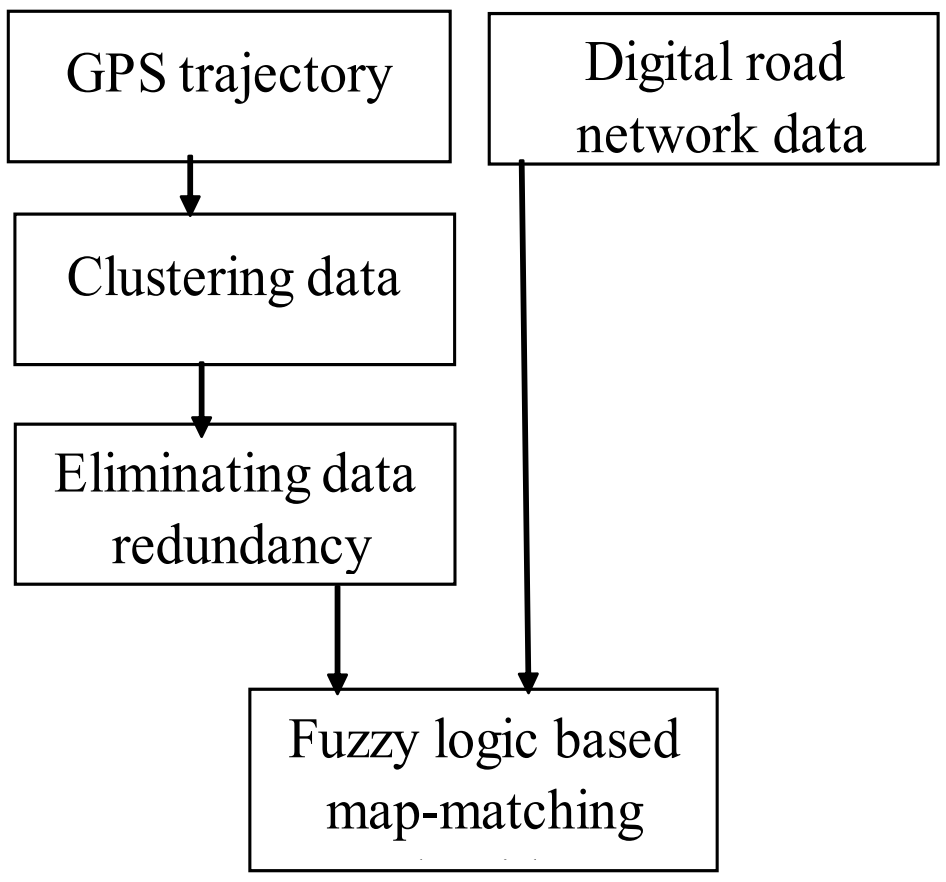

Fig. 12. Proposed method follow-chart

### *C. Clustering stay-point regions and eliminating superfluous points*

As previously mentioned, selecting an appropriate value for epsilon (*Eps*) and the minimum number of points (*MinPts*) in DBSCAN clustering is essential. We prefer to have fewer data in each cluster over losing valuable data necessary for the map-matching algorithm. Therefore, the selection of parameters is based on keeping valuable data.

To omit unneeded data, one point from each cluster represents the cluster, and the remaining points are cleared. The average value of stay-points will calculate the representative point such as longitude and latitude of all stay-points in that cluster. Lastly, we will use fuzzy-logic based map-matching method, described in section II. B, to evaluate our approach.

## IV. RESULTS

In this section, we will evaluate the proposed method in detail. As mentioned earlier, there are two main criteria for assessing map-matching approaches. The first is the processing time of the scheme, and the second is counting the number of correct and incorrect road link matches. In addition, the volume efficiency and average execution time speed for each point are calculated for more evaluation and comparison.

### *A. Result of the DBSCAN clustering on the dataset*

To select the epsilon, we looked at the diagram's behavior from the point of its k-nearest neighbor (K-NN). The value of 3 for the minimum number of points based on the dataset's data density is selected. Hence, we calculated the nearest three neighboring points for data. 3-nearest neighbor (3-NN) distances are arranged in ascending order to plot a curve. Fig. 13. shows this curve.

As it is provided, this dataset had multiple varieties of densities. Each elbow on the curve specified one optimal value for *Eps*. The Eps' optimal value considering not eliminating valuable points existed in the first elbow, which is 0.00002, is presented in a dashed line on Fig. 13.

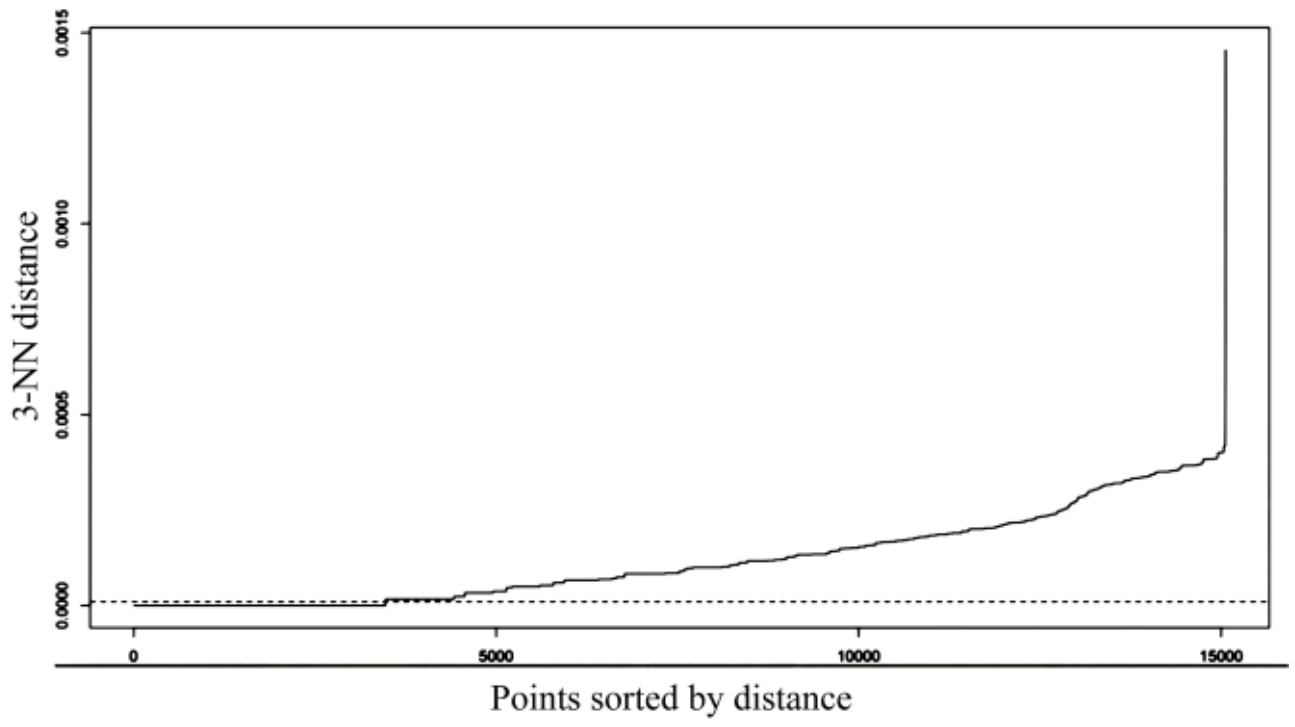

Fig. 13. Three nearest neighbor distance curves for dataset

The DBSCAN algorithm successfully clustered the dataset's stay-point regions. DBSCAN clustering was performed in WEKA version 3.6.9 on 7531 GPS data for 8:24 seconds with a minimum point of 3 and epsilon of 0.00002. The result generated 133 clusters and 5334 non-cluster data. In total, we achieved 5467 new points for the map-matching process. Fig. 14. stated the detailed performance of DBSCAN clustering for alternative elbows on the curve, clarified other noteworthy values of *Eps*.

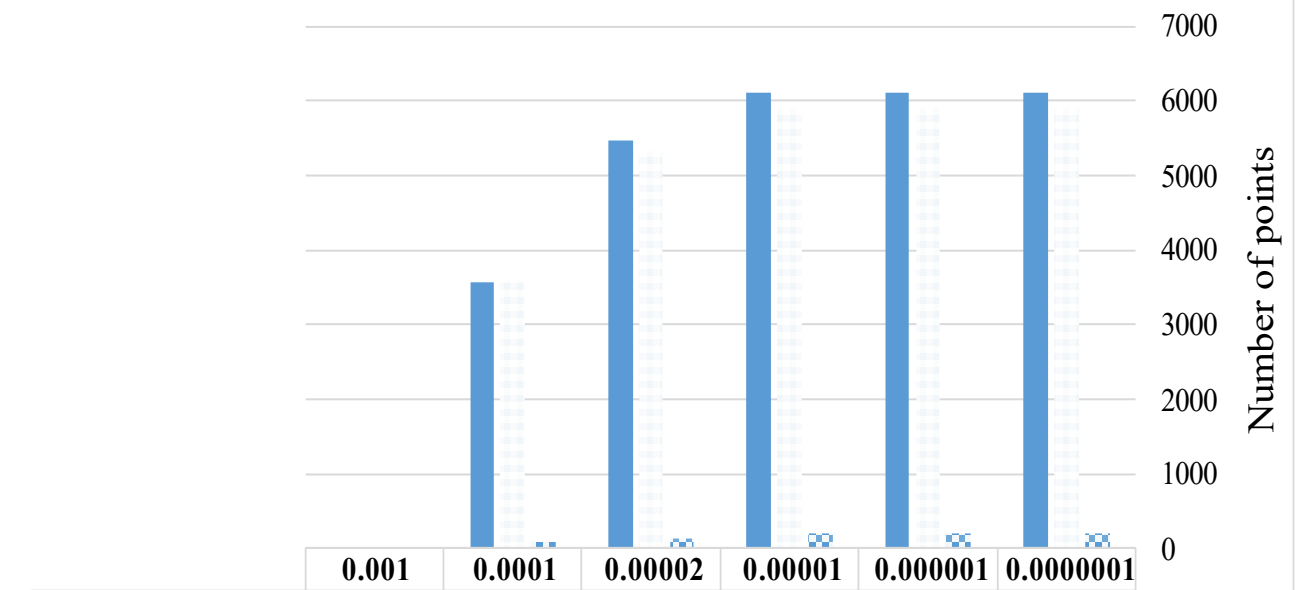

| | 0.001 | 0.0001 | 0.00002 | 0.00001 | 0.000001 | 0.0000001 |
|---|---|---|---|---|---|---|
| Clustered | 4 | 94 | 133 | 195 | 198 | 195 |
| Non-clustered | 1 | 3586 | 5334 | 5910 | 5910 | 5910 |
| Sum of output data | 5 | 3580 | 5467 | 6105 | 6108 | 6105 |

Fig. 14. Number of clustered and non-clustered points for different epsilons

After identifying the stay-point clusters in the dataset, each group is replaced by one point, and the surplus points are removed from the dataset. The replacing data element's attribute equalized the average features of stay-points on that cluster. The efficiency and adequacy of our approach on two parts of the trajectory showed in Fig. 15.

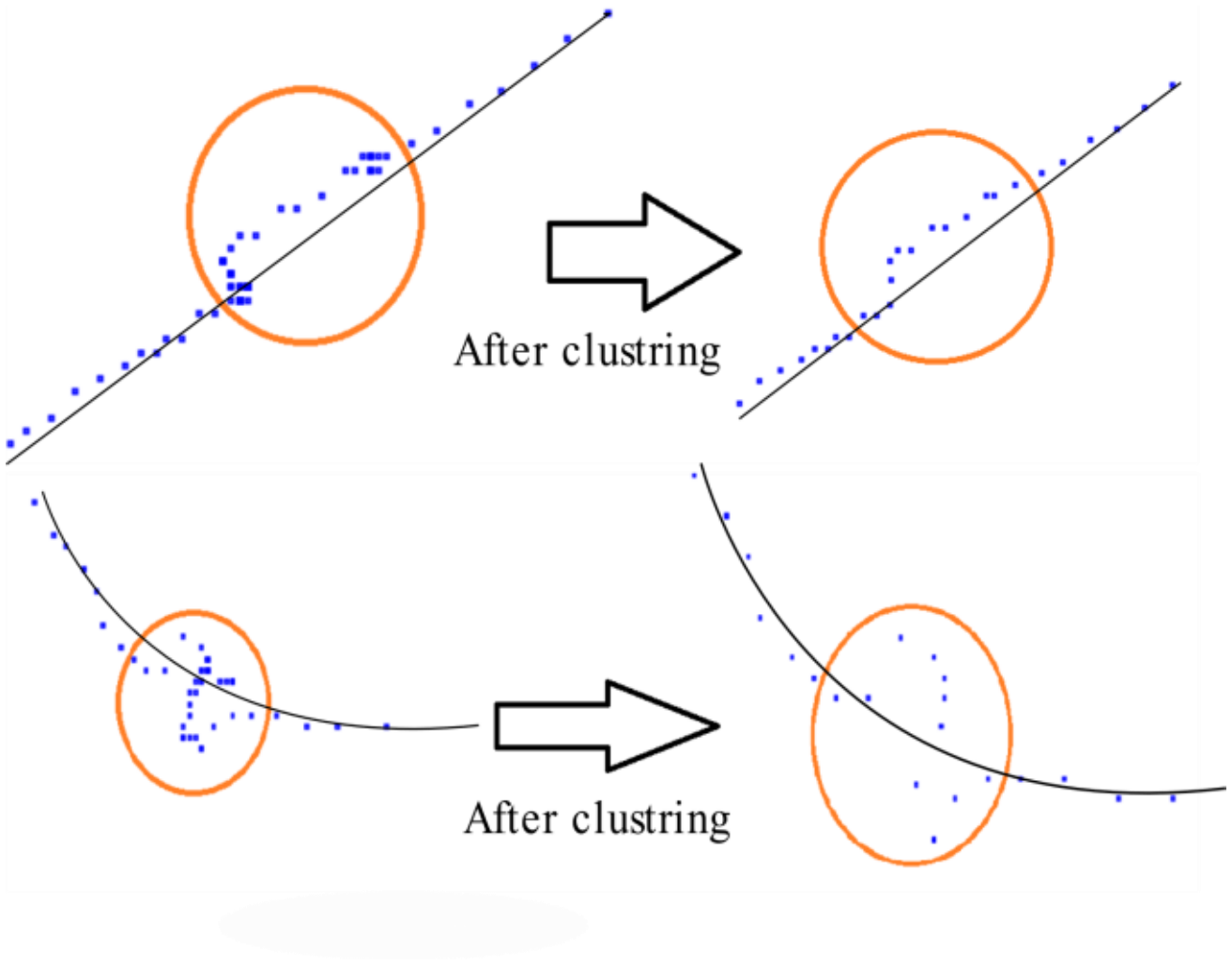

Fig. 15. Clustered stay-points data with DBSCAN

Finally, the fuzzy-logic based map-matching algorithm [8] is implemented on the raw dataset, and the stay-points eliminated dataset to obtain experimental results.

## B. *Result of the fuzzy-logic based map-matching algorithm*

The previous method and approach were implemented by the Rstudio version 3.4 and run on a computer with an Intel Core i5 processor and 16 GB memory in MacOS. Subsequently, both procedures are compared based on four metrics: accuracy, processing time, speed, and volume efficiency.

Both approaches identified the same road path on the ground truth data with the same number of correct link identification. Our method maintained the previous approach's accuracy with more proficiency. Fig. 16. the chart illustrated this subject matter in more detail.

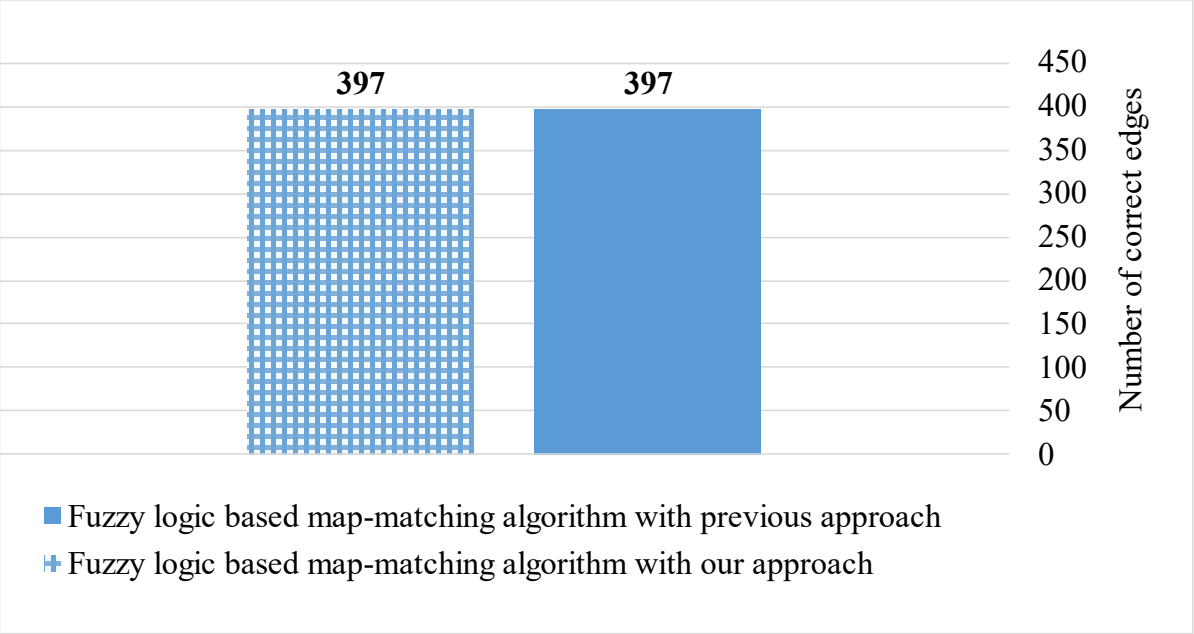

Fig. 16. Accuracy performance evaluation: number of correct links detection among 399 links

The processing time of the map-matching algorithm declined by 8.9 percent—the result of this evaluation among two approaches clarified in Fig. 17.

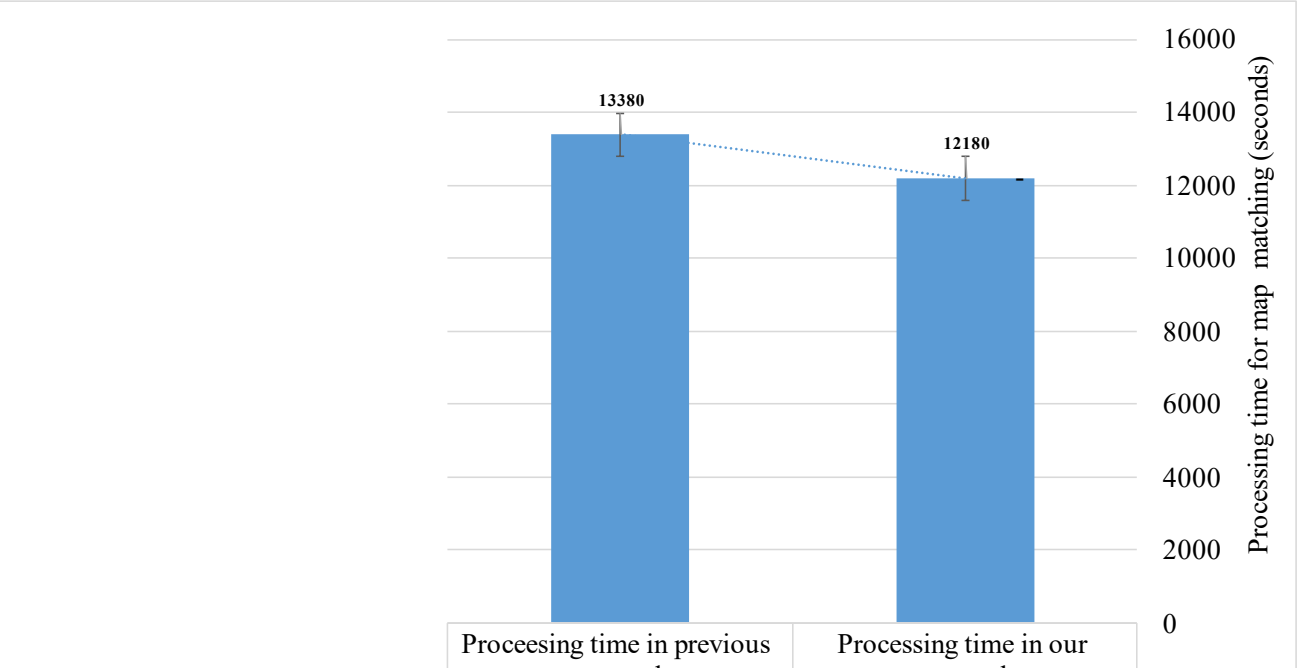

| | Proceesing time in previous approach | Processing time in our approach |
|---|---|---|
| Processing time for map-matching | 13380 | 12180 |

Fig. 17. Time efficiency evaluation: overall processing time comparison between two approaches

The dataset volume decreased by 27.39 percent, significantly impacting time and storage resources. Topic pointed out in Fig.18.

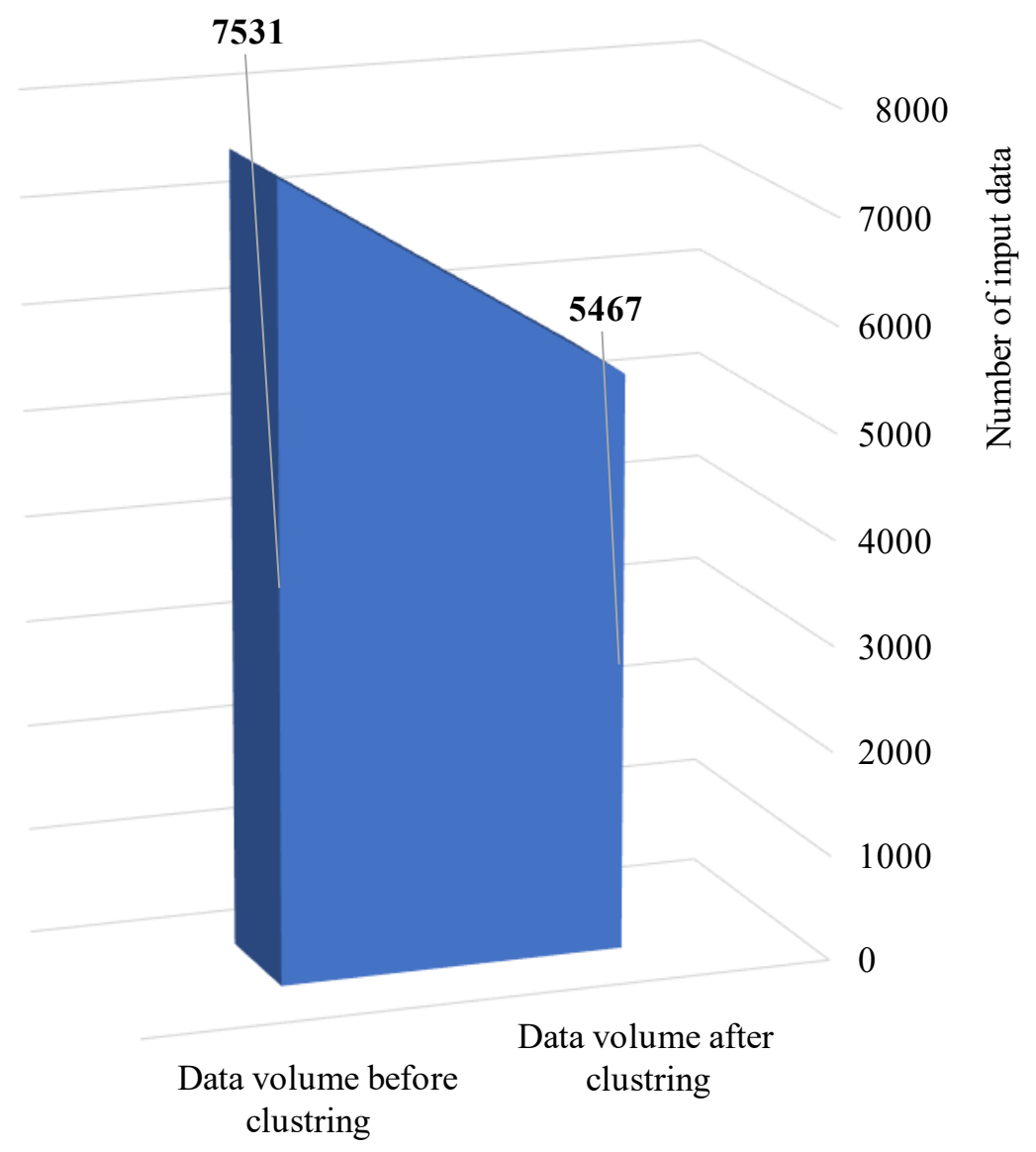

| | Data volume before clustring | Data volume after clustring |
|---|---|---|
| Number of input data | 7531 | 5467 |

Fig. 18. Volume efficiency evaluation: the proposed approach decreased the data volume

Our approach performed a faster execution time by 21.42 percent than the earlier method. Fig. 19. Referred to more elements of this subject.

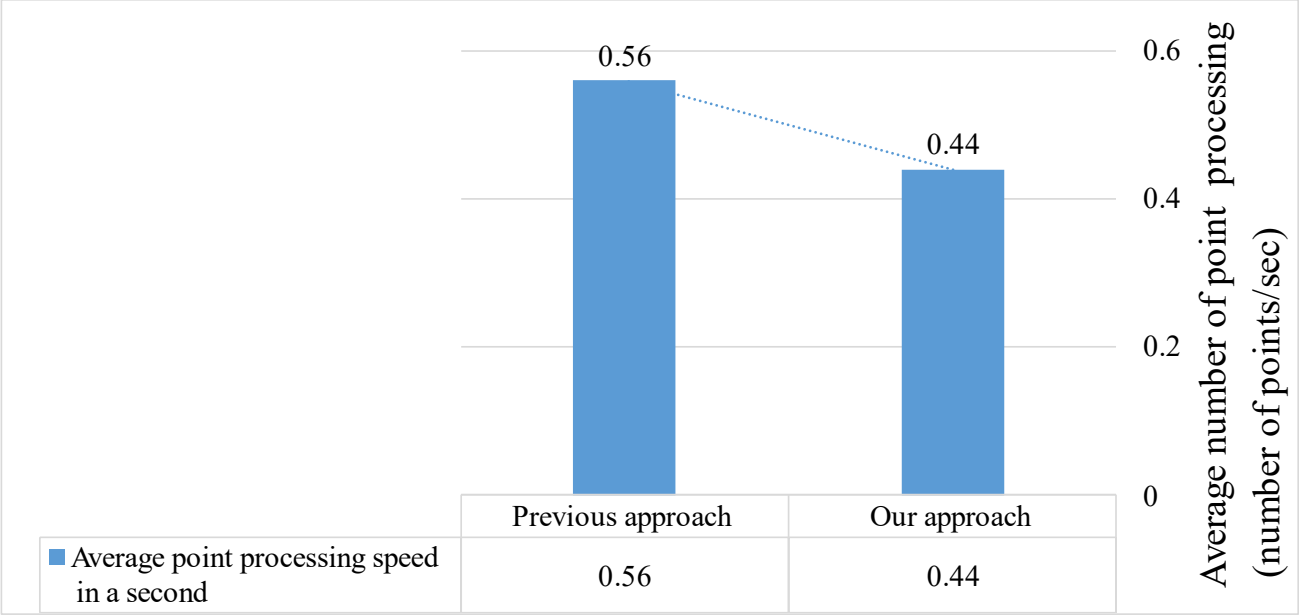

Fig. 19. Speed efficiency evaluation: average execution time speed for each point

The successful outcome of the map-matching algorithm in a small trajectory section is exhibited in Fig. 20. Raw GPS points were displayed as blue round dots, and map-matched spots appeared as red small square. The result indicated how accurately points are matched without confusion in path track or mismatching in the intersection.

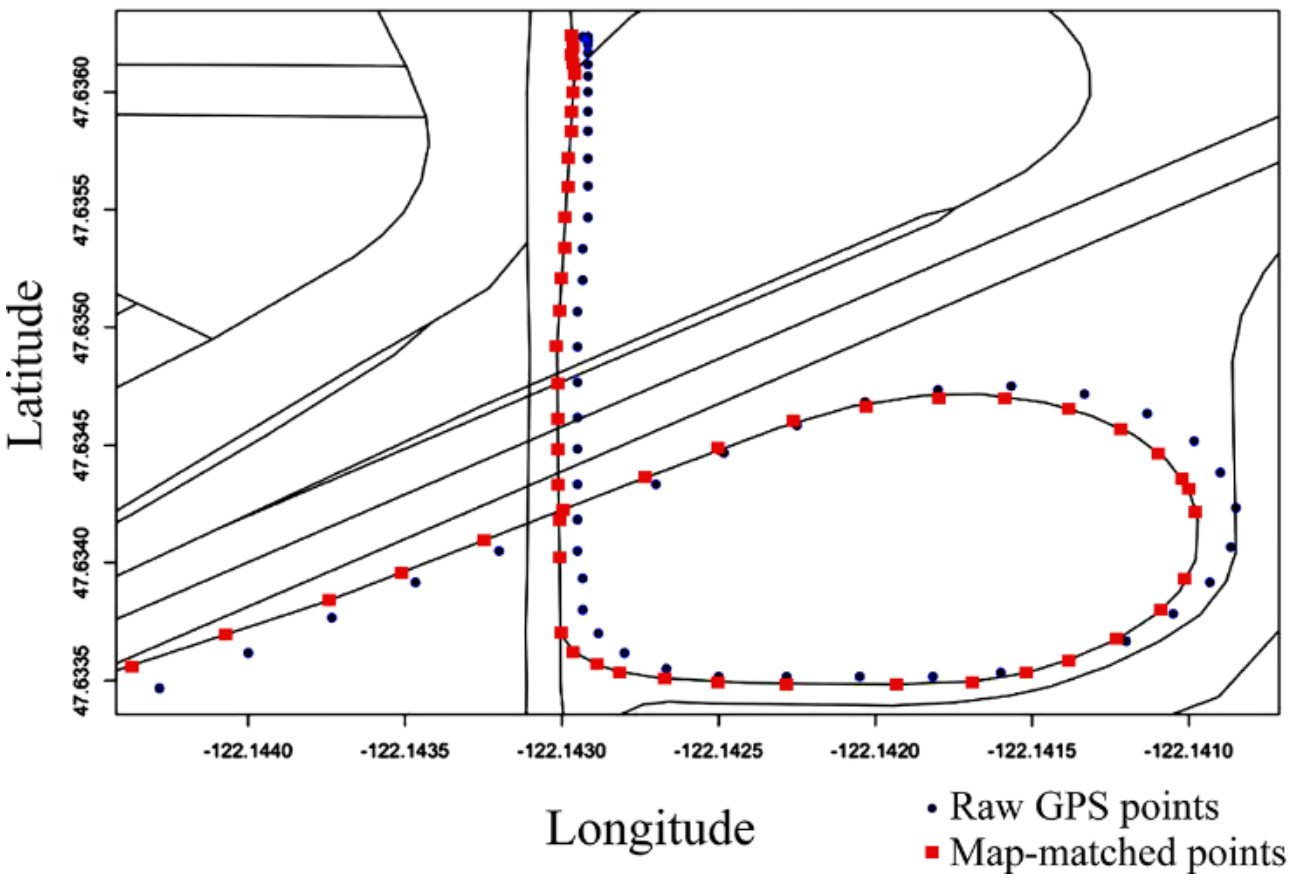

Fig. 20. Map-matching sample on the dataset

## V. Literature Review

A simple map-matching algorithm for car navigation was presented in 1996 by Kim lee [18]. Generally, map-matching methods are divided into online or offline sampling with high or low sampling. This section discusses some of the most important and recently studied researchers.

Works on low-frequency sampling datasets are prevalent. [19] developed their method based on fuzzy logic to identify where the crashes happened. Relation of the crash locations to the road geometry features or the traffic jam on specific areas was analyzed to determine the road network's dangerous parts. [20] used A* shortest path-search-algorithm to obtain the shortest path between two consecutive GPS points. Their method can be categorized as the topological algorithm. They designated a specific weight to each parameter and optimized it using a Genetic Algorithm (GA). [21] developed a novel map-matching algorithm and named it ST-matching. Their algorithm generated a candidate graph using spatial and temporal analysis, where the edges of the chart showed the roads. Then, a sequence of matched points is observed. The path with the highest sum of the matching points' probabilities was identified as the matching result. In [22] developed an if-matching algorithm to map-match their taxi trajectory dataset, sampled in low frequency and unstable situations. Their approach used available data measurement parameters such as speed, direction, and locations. The research [23] introduced a novel map-matching method based on evaluating the curvedness of GPS trajectory. [24] developed a collaborative map matching method (CMM) to solve low sampling rate GPS trajectories' mapping by processing tracks in batches. The authors [34] proposed a map-matching method based on deep learning using advanced spatial-temporal analysis (DST-MM).

High-frequency sampling datasets also face multiple challenges. This method [8] is developed based on fuzzy logic and stands among the most accurate map-matching algorithms. In this work, the mapping algorithm was divided into three steps. It used GPS trajectory features as the inputs of three different fuzzy interface systems. Approach [9] is one of the best topological map-matching algorithms. Despite applying a topological method, their work in identifying the correct link is noticeable. They used two new parameters: The road connectivity parameter and the turn restriction parameter (right turn, left turn, and U-shaped turn). This [25] focused on the uncertainty of GPS sensors and road identification requirements. In this paper, they developed the Kalman filter method. This method's advantage is its' low-cost implementation with a gyroscope and a single-frequency GPS receiver. Some other works implemented on different types of moving objects datasets, such as wheelchair or pedestrian locations in indoor or outdoor environments. This work used advanced map-matching algorithms to recognize their location. [26] introduced fuzzy-logic based map-matching to estimate the location of the wheelchair in the sidewalk network. They used two main criteria (the distance from the GPS point in each segment and direction difference between GPS trajectory and segment) to determine the best segment among candidate links. However, tracking the user's path in this method required three future GPS adaptive points in addition to the current GPS matched point. Moreover, they did not benefit from GPS's additional information due to the low speed of their wheelchair data [18]. [27] firstly, divided trajectories into several segments, secondly, preprocessed the road network, and thirdly, built a multi-layer road index system. lastly, a map-matching strategy determines the best match for each segment. [28] developed a new dynamic two-dimensional (D2D) weight-based map-matching algorithm to consider road widths as an input, primarily neglected in previous works. Other groups [29] focused on the junction-matching problem by the junction decision domain model, which considered the road segments' width, the angle between two roads, and the road network's accuracy and GPS points. They used HMM algorithm to improve map-matching. Furthermore, [33] provided a method to determine high-risk driving events from smartphone sensors employing HMM.

Online map-matching algorithm researchers developed the Spatial-direction-matching (SD-matching) algorithm in [30], a three-stage online map-matching algorithm. Their other work [31] added an online trajectory compression algorithm named Heading Change Compression (HCC) to their previous work to find a brief and compact trajectory representation. [32] presented a new map-matching algorithm based on the floor map to recognize the path segments and the user's paths on indoor locations. This approach used a user's online smartphone GPS trajectory to identify the accurate location of the user.

Map-matching researchers generally argue that advanced map-matching algorithms such as fuzzy-logic based map-matching locate the GPS trajectory more accurately than other algorithms due they often use more inputs and complicated procedures to recognize the correct segment [18]. However, they have a high processing time. This paper aims to close the gap between accuracy and efficiency. Although researchers have proposed many map matching methods, they often fail to balance the two conflicting objectives, i.e., correct link identification and computation time.

## VI. CONCLUSION

Map-matching methods encounter challenges in counterbalancing the accurateness and processing span of the procedure. In this work, stay-points are recognized as one of the causes that map-matching algorithms have increased processing duration.

Our research operated the DBSCAN algorithm to pinpoint the stay-point region of the GPS trajectory and eliminated the redundant points, which reduced the efficiency of the fuzzy-based-map-matching algorithm. We measured our method with the canonized fuzzy-logic based map-matching algorithm over ground truth data in the real world. Subsequently, the map-matching algorithm's processing duration lowered by 8.9 percent. Accordingly, the performance pace for each point accelerates, and execution speed is up by 21.42 percent. Our method can be used as an offline preprocess for moving object database analysis and applications.

We intend to work with diverse datasets that include additional unconventional stay-points and unusual patterns in their trajectory in future work. The ambition is to analyze vehicles' immobile behavior to improve urban planning and management. Car accidents, closed/narrow roads, or slow road traffic flow have distinct irregular patterns, which can be noticed and designed to facilitate transportation.

Similarly, two critical impediments for the map-matching algorithms are pair of subsequent trajectories' inadequacy in critical regions, particularly border points, which occurs predominantly in low-frequency sampling data, and Y-shaped junction. This hardship compels most of the inaccuracies in the link identification methodology. A notable portion of incorrect link matching is the outcome of this problem. We plan to discover an optimal resolution to solve this issue in future works.